\documentclass[10pt,twocolumn,letterpaper]{article}

\usepackage{iccv}
\usepackage{times}
\usepackage{graphicx}
\usepackage{amsmath}
\usepackage{amssymb}
\usepackage{multirow}
\usepackage{algorithm,algorithmic,subfigure,booktabs,ctable}
\usepackage{enumerate}

\usepackage[breaklinks=true,bookmarks=false]{hyperref}

\iccvfinalcopy 

\def\iccvPaperID{****} 

\begin{document}
	
	\title{Transferable Knowledge-Based Multi-Granularity Aggregation Network for Temporal Action Localization: Submission to ActivityNet Challenge 2021}
	
	\author{Haisheng Su$^{1\ast}$, Peiqin Zhuang$^{1,2}$, Yukun Li$^{1}$, Dongliang Wang$^{1}$,  Weihao Gan$^{1}$, Wei Wu$^{1}$, Yu Qiao$^{2,3}$ \\
		\\
		$^{1}$SenseTime Research\\
		$^{2}$SIAT-SenseTime Joint Lab, Shenzhen Institutes of Advanced Technology, Chinese Academy of Sciences\\
		$^{3}$Shanghai AI Laboratory, Shanghai, China \\
		{\tt\small \{suhaisheng,zhuangpeiqin,liyukun,wangdongliang,ganweihao,wuwei\}@sensetime.com, yu.qiao@siat.ac.cn}
		\\
	}
		
\maketitle
\pagestyle{empty}  
\thispagestyle{empty} 
\let\thefootnote\relax\footnotetext{$\ast$ Corresponding author.}

	\begin{abstract}
		This technical report presents an overview of our solution used in the submission to 2021 HACS Temporal Action Localization Challenge on both \textbf{Supervised Learning Track and Weakly-Supervised Learning Track}. Temporal Action Localization (TAL) requires to not only precisely locate the temporal boundaries of action instances, but also accurately classify the untrimmed videos into specific categories. However, Weakly-Supervised TAL indicates locating the action instances using only video-level class labels. In this paper, to train a supervised temporal action localizer, we adopt Temporal Context Aggregation Network (TCANet) to generate high-quality action proposals through ``local and global" temporal context aggregation and complementary as well as progressive boundary refinement. As for the WSTAL, a novel framework is proposed to handle the poor quality of CAS generated by simple classification network, which can only focus on local discriminative parts, rather than locate the entire interval of target actions. Specifically, we propose to utilize convolutional kernels with varied dilation rates to enlarge the receptive fields, which is found to be capable of transferring the discriminative information to surrounding non-discriminative regions. Then we design a cascaded module with proposed Online Adversarial Erasing (OAE) mechanism to further mine more relevant regions of target actions through feeding the erased feature maps of discovered regions back to the system. Besides, inspired by the transfer learning method, we also adopt an additional module to transfer the knowledge from trimmed videos (HACS Clips dataset) to untrimmed videos (HACS Segments dataset), aiming at promoting the classification performance on untrimmed videos. Finally, we employ a boundary regression module embedded with Outer-Inner-Contrastive (OIC) loss to automatically predict the boundaries based on the enhanced CAS. Our proposed scheme achieves \textbf{39.91} and \textbf{29.78} average mAP on the challenge testing set of supervised and weakly-supervised temporal action localization track respectively.
	\end{abstract}
	
	\section{Introduction}

With increasing development of computer vision as well as the great amount of media resources, intelligent video content analysis has attracted much attention from many researchers in recent years. Videos in realistic life are usually long and untrimmed, which may contain multiple action instances with arbitrary durations. It leads to an important yet challenging task for video analysis: Temporal Action Localization (TAL), which requires to not only classify the untrimmed videos into specific categories accurately, but also locate the temporal boundaries of action instances precisely. Although substantial progress has been achieved on this task \cite{zhou2015learning},\cite{SCNN}, \cite{SSN},\cite{SSAD},\cite{CBR},\cite{BSN},\cite{Hao2018Temporal},\cite{Guo2018Fully}, it is still limited for industrial applications due to the huge amount of temporal annotations used for training such a deep learning based model in a fully-supervised manner, which are labor-intensive to annotate especially for a large-scale dataset. On the contrary, weak labels such as video-level labels are much easier to obtain, hence many current works try to handle this problem under weak supervision.

\begin{figure*}[t]
\centering
\includegraphics[width=14.7cm, height=2.5cm]{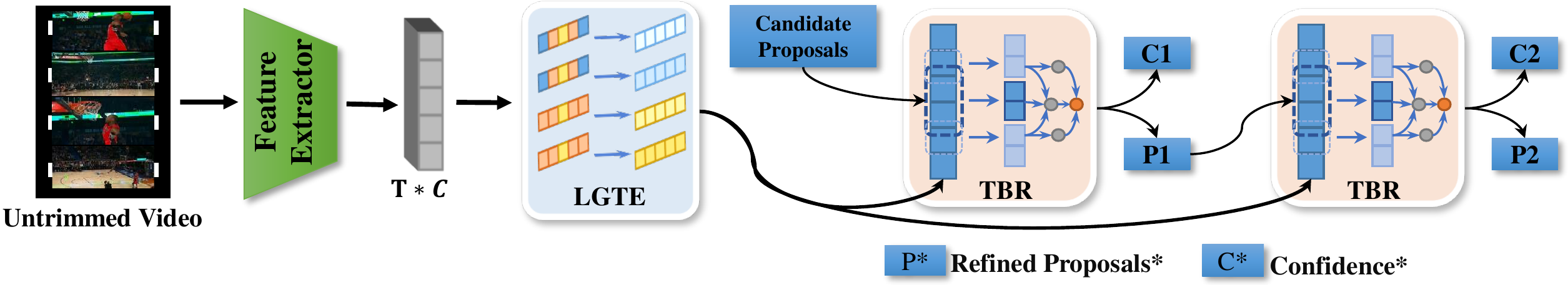}
\caption{{\color{black}The framework of TCANet. TCANet mainly contains two modules: LGTE and TBR.
LGTE is employed to capture \textit{local-global} temporal inter-dependencies simultaneously. TBR is adopted to perform frame-level and segment-level boundary regressions, respectively.
Finally, high-quality proposals are obtained through complementary fusion and progressive boundary refinements.
}
}
\label{fig_framework}
\vspace{-0.2cm}
\end{figure*}

Most existing supervised TAL methods follow a two-stage scheme~\cite{wang2015ssn,singh2016anet_winner}, namely temporal action proposal generation and classification. Although action recognition methods~\cite{wang2016tsn,feichtenhofer2019slowfast} have achieved impressive classification accuracy, the TAL performance is still unsatisfactory in several mainstream benchmarks~\cite{thumos14, caba2015activitynet, zhao2019hacs}. Hence, many researchers target improving the quality of temporal action proposals. In this technical report, we adopt Temporal Context Aggregation Network (TCANet) for high-quality proposal refinement, as shown in Figure~\ref{fig_framework}. First, the Local-Global Temporal Encoder (LGTE) is proposed to simultaneously capture \textit{local and global} temporal relationships in a channel grouping fashion, which contains two main sub-modules. Specifically, the input features after linear transformation are equally divided into $N$ groups along the channel dimension. Then Local Temporal Encoder (LTE) is designed to handle the first $A$ groups for local temporal modeling. At the same time, the remaining $N - A$ groups are captured by the Global Temporal Encoder (GTE) for global information perception. In this way, LGTE is expected to integrate the long-term context of proposals by global groups while recovering more structure and detailed information by local groups. Second, the Temporal Boundary Regressor (TBR) is proposed to exploit both boundary context and internal context of proposals for frame-level and segment-level boundary regressions, respectively. Finally, high-quality proposals are obtained through complementary fusion and progressive boundary refinements.

Analogous to Weakly Supervised Object Detection (WSOD) in images, Weakly Supervised Temporal Action Localization (WSTAL) can be regarded as a temporal version of WSOD which aims to locate the action instances using only video-level class labels. However, WSTAL is much more challenging compared to WSOD as a result of the larger video content variation and uncertain temporal length of action instances. A prevalent practice of WSTAL adopts the idea \cite{B.Zhou} to generate the 1-D Class Activation Sequence (CAS) to highlight the discriminative regions contributing to the video classification results most, which is originally used for locating the object in images. Nevertheless, a high-quality CAS used for temporally locating the action boundaries should possess the following two properties: (1) CAS can completely cover the temporal interval of target actions; (2) CAS can densely locate the action instances with less missing detections.
	
To effectively locate the action instances under weak supervision, we propose the Multi-Granularity Fusion Network (MGFN), which firstly adopts a cascaded dilated classification block to enhance the quality of CAS, then employs a boundary regression module to directly predict the temporal boundaries of action instances based on the CAS. In order to generate a high-quality CAS, our cascaded dilated classification block utilizes two main sub-modules to achieve this goal. Specifically,  multi-dilated convolution module augments the simple classification network with multiple convolutional kernels of different dilation rates, on the purpose of transferring the discriminative information of initial seeds to surrounding non-discriminative regions, thus to expand the visible areas of classification network. Then the cascaded classification module adapts two classifiers with the same architecture to further discover other potential action regions using an Online Adversarial Erasing (OAE) mechanism, which do not appear on the initial localization sequence. With this mechanism, the feature maps of discriminative regions discovered in the first stage are dynamically erased, which are then fed to the second-stage classifier for further mining. Unlike previous methods, our approach is more efficient and intuitive. Besides, in order to further improve the CAS quality, we incorporate the transfer learning idea and learn transferable knowledge between trimmed videos and untrimmed videos, aiming at promoting the classification performance on untrimmed videos. Finally, instead of performing temporal action detection via thresholding the CAS directly, which may not be robust to noises in CAS, we adopt a boundary regression module to predict the boundaries. As for the segment-level supervision used for boundary regression, the Outer-Inner-Contrastive (OIC) loss is employed. The whole framework is optimized in an end-to-end fashion.
	
It is non-trivial to enhance the quality of CAS effectively and efficiently, since the CAS generated by simple classification network can only focus on local discriminative parts, which is inferior and is not qualified for TAL task. Hence, so as to achieve a good performance on this task, a high-quality CAS is the prerequisite. For the sake of end-to-end optimization, our OAE mechanism is time-efficient and only needs to train a model for entire regions mining. Therefore, with the integrated training process, the augmented classification network based on transferable knowledge together with the boundary regression module can collaborate with each other for a better performance.

\section{TCANet - Supervised Learning Track}

 As shown in Figure~\ref{fig_framework}, we propose \textbf{Temporal Context Aggregation Network (TCANet)} to generate high-quality proposals, which mainly consists of two main modules: Local-Global Temporal Encoder and Temporal Boundary Regressor.
Firstly, the Local-Global Temporal Encoder (LGTE) is adopted to simultaneously encode the input video features' \textit{local and global} temporal relationships. {\color{black} Then the Temporal Boundary Regressor (TBR) is utilized to refine the boundaries of the proposals by exploiting both boundary and internal context for frame-level and segment-level boundary regressions, respectively.}


\begin{figure}[t]
\centering
\includegraphics[width=8.2cm,height=6.0cm]{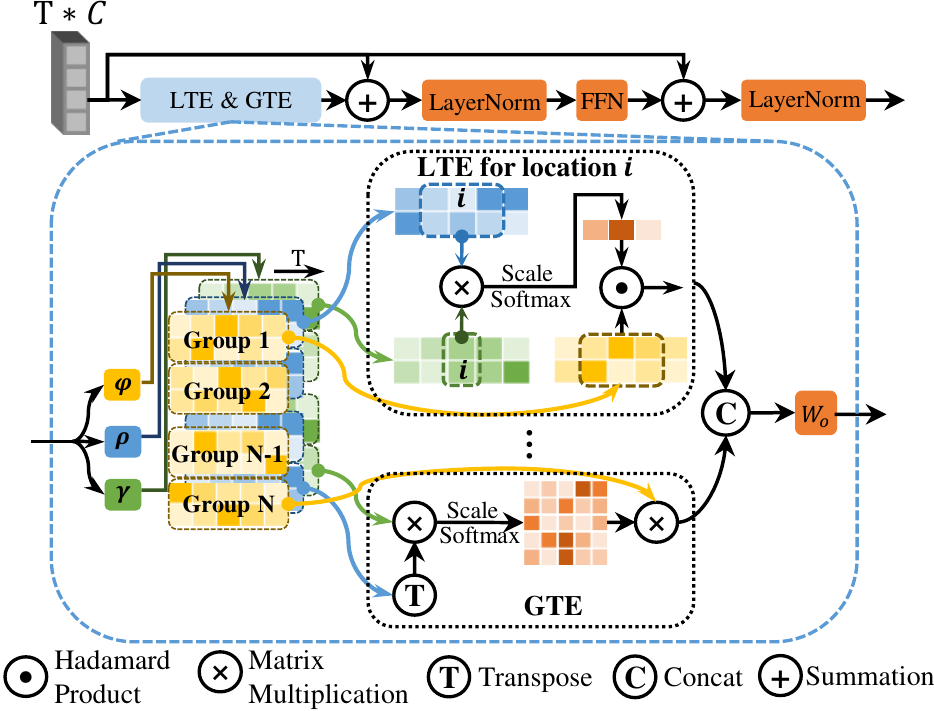}
\caption{The detailed structure of Local-Global Temporal Encoder (LGTE).}
\label{fig_lgte} 
\vspace{-0.2cm}
\end{figure}

\subsection{Local-Global Temporal Encoder}
For long videos, long-term temporal dependency modeling is essential, proven by many previous works~\cite{gao2020rapnet,wu2019lfb}. 
Nonlocal~\cite{wang2018nonlocal} is often applied to obtain the relationship between different global locations.
However, global modeling only is easy to introduce global noise and insensitive to small boundary changes. 
We propose a local and global joint modeling strategy to alleviate this problem, as shown in Figure~\ref{fig_lgte}.

\noindent \textbf{Local Temporal Encoder (LTE)} is responsible for capturing local dependencies based on local details dynamically. Precisely, to measure the relationship between temporal location $i$ and its local areas, the $cosine$ similarity between two temporal locations is adopted to generate  similarity vector $S_i^l$ and weight vector $W_i^l$:
\begin{equation}
    S_i^l=\gamma^l(f_{i})\cdot(\rho^l([(f_{i-\lfloor w/2 \rfloor})^T,\cdot \cdot \cdot (f_{i+\lfloor w/2 \rfloor})^T]^T))^T \in R^{1\times w}
\label{local_sim}
\end{equation}
\begin{equation}
    W_i^l = \text{Softmax}(\frac{S_i^l}{\sqrt{C}})
\label{softmax_fuse1}
\end{equation}
where $C$ is the number of channels, $w$ is the size of the modeling area for location $i$, which is defined as $WindowSize$. For example, the value of $w$ in Figure~\ref{fig_lgte} LTE is 3. $\gamma^l$ and $\rho^l$ are two different linear projection functions that map the input feature vectors to the similarity measure space.

With equation~\ref{local_sim}, the relationship between each location and its corresponding modeling area can be calculated. To achieve local information exchange, the following formula will be utilized to collect local context information from the corresponding local area dynamically:
\begin{equation}
    f_i^{l}=W_i^l \cdot(\varphi^l([(f_{i-\lfloor w/2 \rfloor})^T,\cdot \cdot \cdot (f_{i+\lfloor w/2 \rfloor})^T]^T)),
\label{local_fo}
\end{equation}
where $f_i^{l}$ represents the new expression of location $i$, and $\varphi^l$ is a linear projection function.

\noindent \textbf{Global Temporal Encoder (GTE)} is designed to model the long-term temporal dependencies of videos. Compared with LTE, GTE needs to aggregate global interactions for each location on the temporal dimension. Therefore, the relationship between each location and the global feature is written as follows:
\begin{equation}
    S_i^g=\gamma^g(f_{i})\cdot(\rho^g(F))^T \in R^{1\times T},
\label{global_sim}
\end{equation}
\begin{equation}
    W_i^g = \text{Softmax}(\frac{S_i^g}{\sqrt{C}}),
\label{softmax_fuse2}
\end{equation}
where $\gamma^g$ and $\rho^g$ are two different linear projection functions.
The global interaction feature of location $i$ can be updated by weight vector $W_i^g$:
\begin{equation}
    f_i^{g}=W_i^g \cdot(\varphi^g(F)),
\label{global_fo}
\end{equation}
where $f_i^{g}$ represents the new global feature representation of location $i$, and $\varphi^g$ is a linear projection function.

\noindent \textbf{Local-Global Temporal Encoder (LGTE).} Each location in the video feature sequence can be modeled locally and globally by LTE and GTE, respectively.
However, it is inefficient to combine them in the form of \textit{``LTE-GTE"} simply.
To solve this problem, LGTE is implemented in a channel grouping fashion.
Specifically, as shown in Figure~\ref{fig_lgte}, the input feature is first projected by $\gamma$, $\rho$, and $\varphi$. These outputs are then divided into $N$ groups along the channel dimension. Hence the channel number of each group is $C/N$. The first $A$ groups are handled by LTEs, while the other $N-A$ groups are fed to GTEs. For location $i$, the combined output of local and global features can be written as:
\begin{gather}
    f_{i}^{a}=[(f_{1i}^{l})^T, \cdot \cdot \cdot  (f_{Ai}^{l})^T,  (f_{(A+1)i}^{g})^T  \cdot \cdot \cdot  (f_{Ni}^{g})^T]^T\cdot W_{o},  \label{output_fuse}\\
    f_{i}^{b} = \text{LayerNorm}(f_{i}^{a}) + f_{i}^{a}, \\
    f_{i}^{'} = \text{LayerNorm}(\text{FFN}(f_{i}^{b}) + f_{i}^{b}),
\end{gather}
where $W_{o}$ is a learnable parameter matrix. Inspired by Transformer~\cite{vaswani2017transformer}, FFN is adopted to capture the interaction of features among different groups at $i$-th temporal location: $\text{FFN}(x)=\text{ReLU}(x\cdot W_1+b_1)\cdot W_2+b_2$.

\begin{figure}[t]
\centering
\includegraphics[width=6cm]{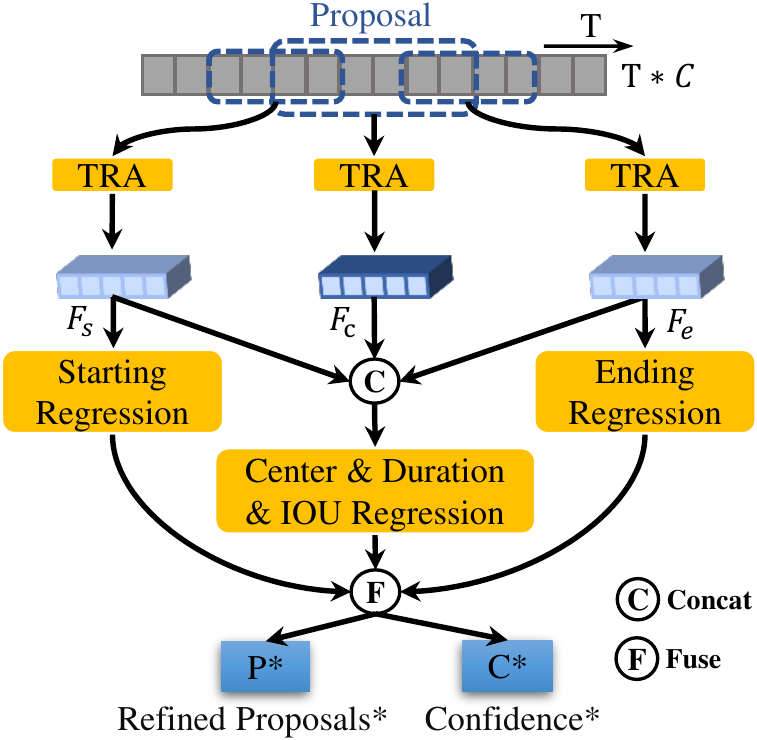}
\caption{The detailed structure of Temporal Boundary Regressor.}
\label{fig_tbr} 
\vspace{-0.2cm}
\end{figure}

\noindent \subsection{Temporal Boundary Regressor}
\textit{Anchor-based methods} ~\cite{shou2016scnn,gao2017turn,lin2017ssad,gao2018ctap,chao2018tal_net,liu2019mgg,gao2020rapnet} leverage the internal context of proposals to regress center location and duration, which can obtain reliable scores but with lower recall.
\textit{Boundary-based methods}~\cite{lin2018bsn,lin2019bmn} only utilize local boundary context to locate the boundaries, which are sensitive to boundaries but with inferior confidence.
{\color{black} Therefore, we propose to combine the boundary context-based frame-level regression and the internal context-based segment-level regression to refine the boundaries.}

\noindent
\textbf{Complementary Regression Strategy.}
As shown in Figure~\ref{fig_tbr}, the feature of one proposal is divided into three parts: the starting context $F_s$, the internal context $F_c$, and the ending context $F_e$. 
To achieve frame-level regression, $F_s$ and $F_e$ are utilized to regress the boundary offset $\Delta \hat{s}$ and $\Delta \hat{e}$ of the starting time and ending time, respectively:
\begin{gather}
     \{\Delta\hat{s}, \Delta\hat{e}\}=\text{Conv1d}(\text{ReLU}(\text{Conv1d}(\{F_s, F_e\})))
\end{gather}
The boundary offsets $\Delta\hat{s}$ and $\Delta\hat{e}$ obtained by this method only utilize the starting and ending local features of proposals. It can effectively reduce noise interference and is more sensitive to the boundary position.

However, only using the local features of the boundary will lose the global context of proposals. Therefore, $F_s$, $F_c$ and $F_e$ are utilized to achieve segment-level regression, which jointly regress the center location offset $\Delta \hat{x}$ and duration offset $\Delta \hat{w}$ of the proposals:
\begin{gather}
    F_a=[F_s, F_c, F_e], \\ 
    \{\Delta \hat{x}, \Delta \hat{w}, p_{conf}\}=\text{Conv1d}(\text{ReLU} (\text{Conv1d}(F_a))), 
\end{gather}
By means of $\Delta\hat{s}$, $\Delta\hat{e}$, $\Delta\hat{x}$ and $\Delta\hat{w}$, two new proposals $(\hat{s_1}, \hat{e_1})$ and $(\hat{s_2}, \hat{e_2})$ can be obtained:
\begin{gather}
    \hat{s_1} = s_p - \Delta \hat{s} w_p, \quad  \hat{e_1} = e_p - \Delta \hat{e} w_p, \\
    \hat{x_2} = x_p - \Delta \hat{x} w_p, \quad  \hat{w_2} = w_p e^{\Delta \hat{w}}, \\
    \hat{s_2} = \hat{x_2} - \hat{w_2}/2, \quad \hat{e_2} = \hat{x_2} + \hat{w_2}/2,
\end{gather}
where $w_p=e_p-s_p$, denotes the length of the proposals.
Finally, the two new proposals will be fused as the final proposals prediction of TBR:
\begin{equation}
    \hat{s} = \tau\hat{s_1} + (1-\tau)\hat{s_2}, \quad \hat{e} = \tau\hat{e_1} + (1-\tau)\hat{e_2},
\end{equation}
where $\tau$ is a fusion parameter, we set it to 0.5 empirically.

%

\noindent \textbf{Progressive Refinement.}
To achieve more accurate boundaries of candidate proposals, a progressive refinement strategy is adopted to generate high-quality proposals from coarse-to-fine. In ablation experiments, we will explore the impact of the number of TBRs on proposal performance.

For more details about the training and inference of TCANet, please refer to our previous published work~\cite{qing2021temporal}.
\section{MGFN - Weakly Supervised Learning Track}
	\begin{figure*}[t]
		\centering
		\includegraphics[height=9.2cm]{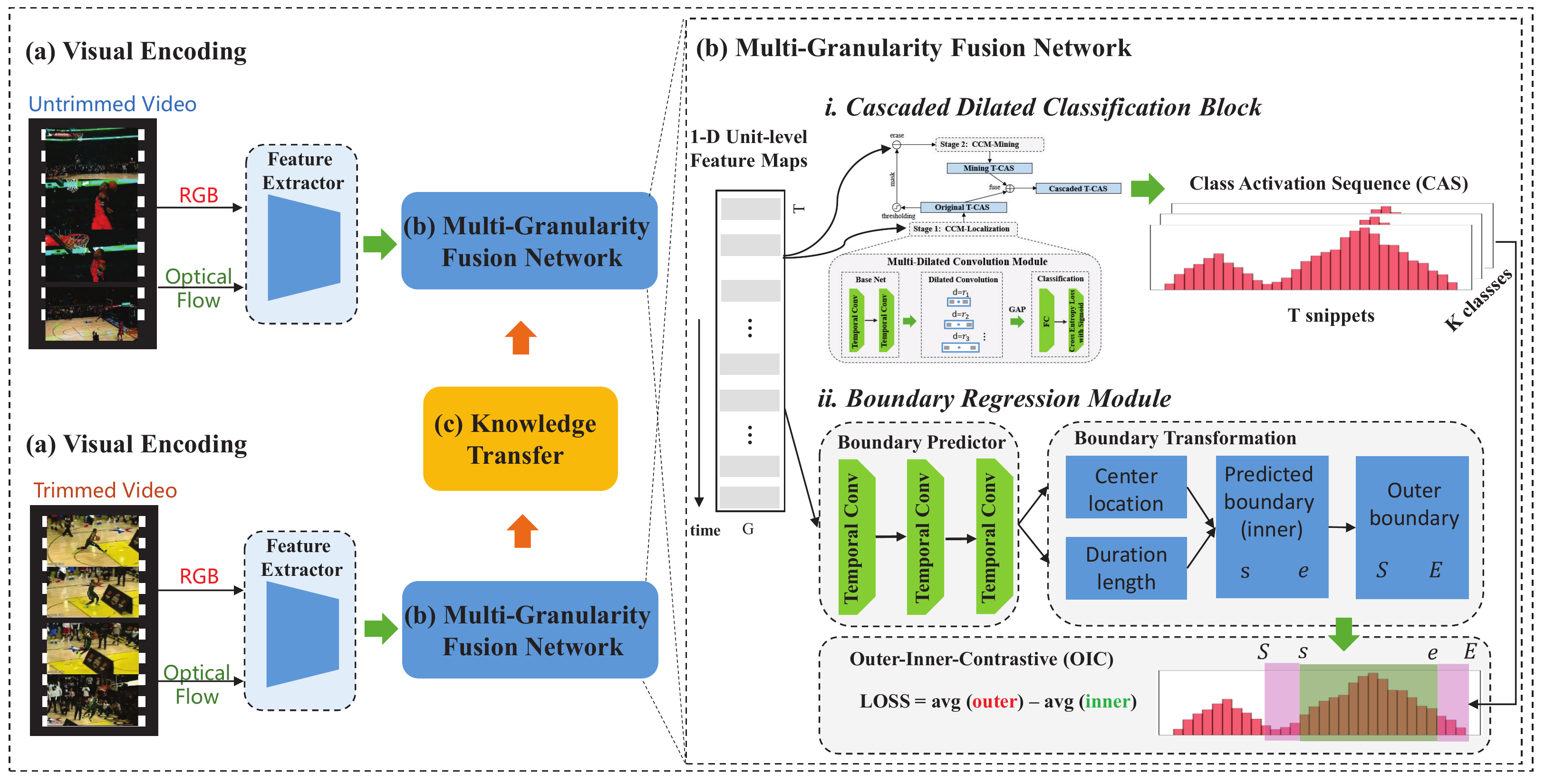}
		\caption{The framework of our proposed WSTAL method. (a) Two-stream network is used to encode video features in snippet-level for our algorithm to perform action recognition and temporal boundary prediction under weak supervision. (b) The architecture of Multi-Granularity Fusion Network: (i) \textit{cascaded dilated classification block} handles the extracted visual features as input for video classification by adopting the multi-dilated convolution module and the cascaded classification module to discover entire class-specific temporal regions; (ii) \textit{boundary regression module} handles the input features to predict the boundary directly, embedded with the Outer-Inner-Contrastive loss to optimize the boundary regressor based on the generated CAS. (c) Finally, we transfer knowledge from trimmed videos to untrimmed videos on the purpose of promoting the classification performance on untrimmed videos, thus to further enhance the quality of Class Activation Sequence (CAS).}
		\label{pic:MGFN}
	\end{figure*}

In this section, we will introduce the technical details of our proposed algorithm, including model architecture, integrated training method and inference strategy. The framework is shown in Fig.\ref{pic:MGFN}.	
	\subsection{Multi-Granularity Fusion Network}
	\label{ssec:mgfn}
	In order to effectively detect the action instances with entire temporal regions and accurate boundaries under weak supervision, we propose a novel architecture to achieve this goal. As shown in Fig.\ref{pic:MGFN}, we design a cascaded dilated classification block so as to mine more relevant regions of target actions through implementing convolutional kernels of varied dilation rates and cascaded mechanism, thus to enhance the quality of Class Activation Sequence (CAS). Instead of locating temporal action instances by directly applying a simple thresholding method, which is not robust to noises of CAS, we adopt a boundary regression module to automatically regress to the accurate boundaries with the help of Outer-Inner-Contrastive (OIC) loss. Besides, since the classification performance of pre-trained video classifier is bound to decrease on untrimmed videos due to the existing of background noises, we introduce the transfer learning mechanism and learn transferable knowledge between trimmed videos and untrimmed videos in order to promote the performance of classification network.

\begin{figure}[!tb]
		\centering
		\includegraphics[height=5.5cm]{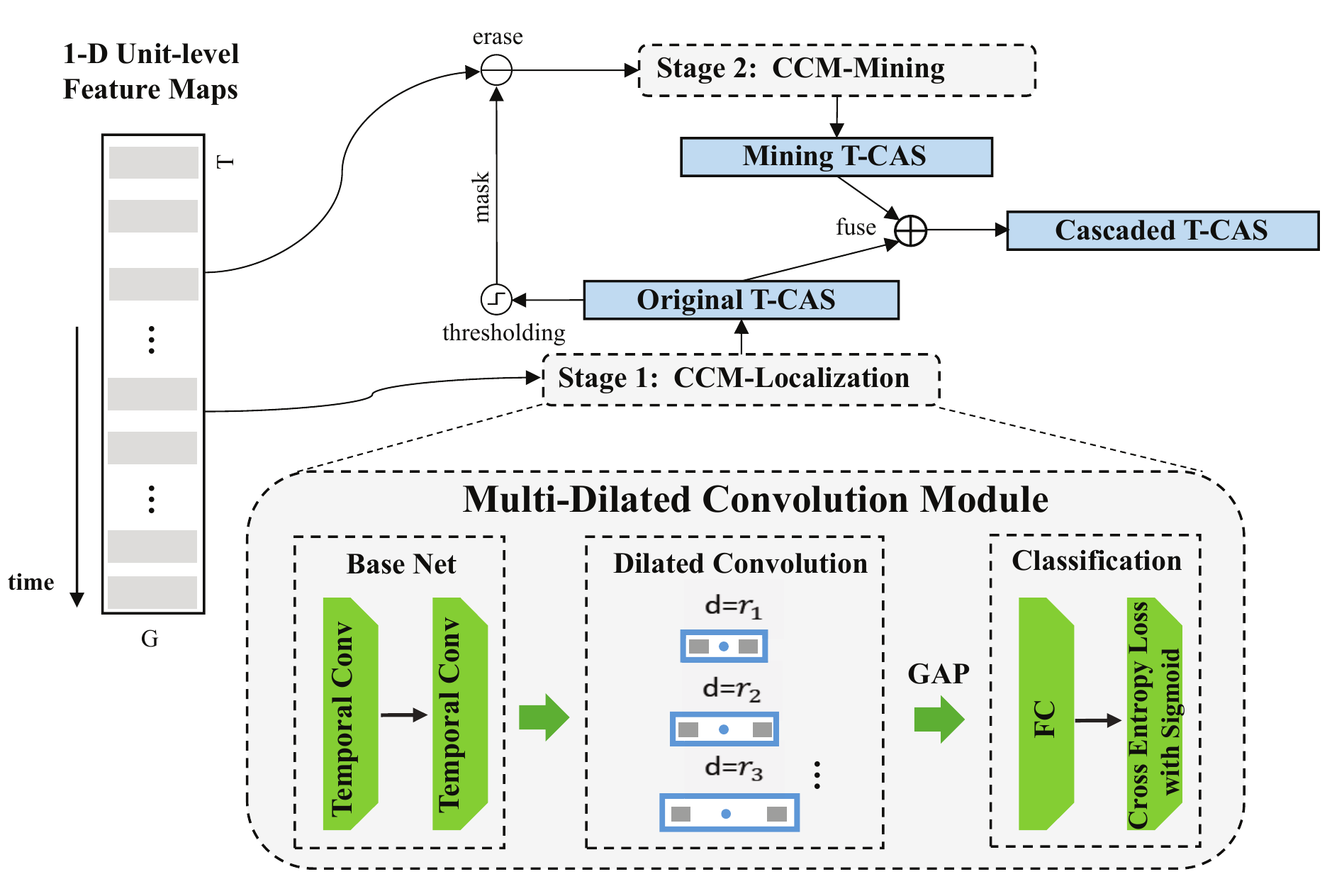}
		\caption{Architecture of the cascaded dilated classification block. The extracted unit-level video features are fed to the two-stage classifiers for video classification concurrently. For each stage, a standard classification network equipped with multiple dilated convolutional branches of varied dilation rates is used for generating a dense localization sequence. Then the two-stage classifiers are combined using an online adversarial erasing mechanism, where the video features of the discriminative regions highlighted in the first-stage localization sequence are dynamically erased for the second-stage classifier, on the purpose of prompting the second classifier to leverage other related regions of target actions. Finally, the localization sequences generated in the two stages are fused for a better quality.}
		\label{pic:cdm}
	\end{figure}

\noindent	{\bf Network Architecture.} The architecture of our MGFN is illustrated in Fig.\ref{pic:MGFN}, which mainly contains two parts: cascaded dilated classification block and boundary regressor. As shown in Fig.\ref{pic:cdm}, the cascaded dilated classification block includes two sub-modules, namely multi-dilated convolution module and cascaded classification module. \textit{Multi-dilated convolution module} is designed to augment the simple classification network by enlarging the receptive field of kernels gradually with multi-dilated convolution branches, which can effectively incorporate the surrounding context and transfer the semantic information from discriminative regions to non-discriminative regions, thus to expand the highlighted areas related to the actions. Then \textit{cascaded classification module} is a two-stage model which combines the two multi-dilated convolution modules of the same structure with an online adversarial erasing method, aiming at discovering more relevant regions of target actions and generate the CAS of high quality. Based on the enhanced CAS, we adopt a \textit{boundary regression module} to predict the segment boundary directly via anchor mechanism, then we inflate the inner segment boundary to obtain the outer segment boundary and optimize the regressor with OIC loss on the purpose of providing a segment-level supervision.
	
\begin{figure}[!tb]
		\centering
		\includegraphics[width=0.99\linewidth]{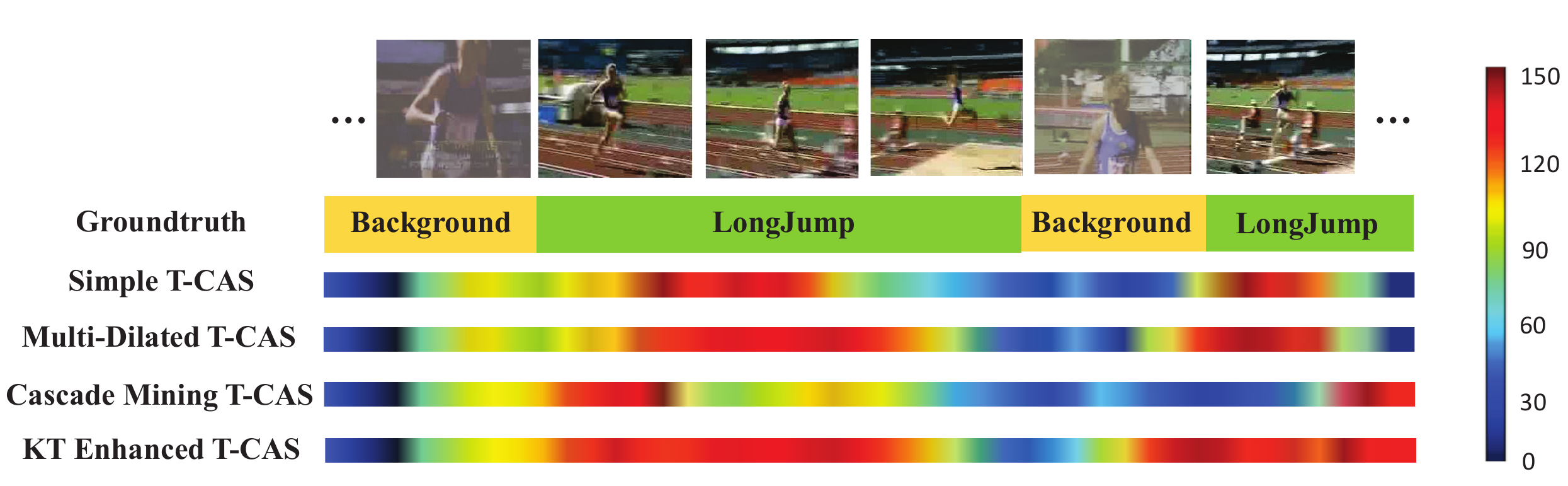}
		\caption{Illustration of the comparison of ground-truth temporal intervals, simple T-CAS, multi-dilated T-CAS, cascade mining T-CAS and knowledge transfer enhanced T-CAS for the \textit{LongJump} action class.}
		\label{pic:ktm}
	\end{figure}

\noindent	{\bf Multi-Dilated Convolution Module.} The goal of this module is to augment the simple classification network with dilated convolution. Since the Class Activation Sequence (CAS) generated by a simple classification network can only highlight discriminative regions of target actions, which is not qualified for temporal action detection task. We introduce the dilated convolution which is already found to be promising in incorporating surrounding context. Through enlarging the receptive field of convolutional kernels with varied dilation rates, semantic information used for supporting the classification result can be transferred from the initially discriminative regions to other surrounding regions, thus to get their discriminativeness enhanced. Fig.\ref{pic:ktm} illustrates how dilation step enables the information to transfer along with the temporal dimension.
	
	Specifically, firstly we adopt two temporal convolutional layers served as base net to handle the temporal information of input video feature sequence, then multi-dilated convolutional branches with varied dilation rates (\textit{i.e. $ d = r_{i}, i = 1, \cdots, k $ }) are appended upon the base net respectively to discover action-relevant temporal regions perceived by varied receptive fields. After Global Average Pooling (GAP) layer, the pooled representations are further passed through a fully connected layer for classification. Then we optimize the augmented classification network with Sigmoid cross-entropy loss, and produce the class-specific localization sequence for each branch respectively. Finally, we obtain the dense Class Activation Sequence (CAS) by fusing the localization sequence from multiple branches. 
	
	Specifically, our dilated convolution module mainly includes two kinds of operation. 1) Standard convolutional kernels with dilation rate $ d=1 $ are employed to generate the original localization sequence $ H_{0} $, where discriminate regions are effectively highlighted in despite of some missing true positive regions. 2) Convolutional kernels with varied dilation rates are employed to expand the discriminative information to surrounding areas. However, we observe that when the receptive field of kernels is set too large, it would also introduce some true negative temporal regions. Hence we choose the small dilation rates (\textit{i.e. d = 2, 3, 5}) in this paper. The final localization sequence $ H $ for temporal action region generation is then fused with $ \mathbf{H} = \mathbf{H}_{0} + \frac{1}{n_{d}}\sum_{i=1}^{n_{d}}\mathbf{H}_{i} $, where $ n_{d} $ is the number of dilated convolution branches.
	
\noindent	{\bf Cascaded Classification Module.} This module aims to further mine more relevant regions of target actions. Although dilated convolution can be used to expand the initially discriminative regions to be more integral, it fails to locate other regions of interest which do not appear on the initial localization sequence. In order to further promote the quality of CAS, we adopt a cascaded mechanism with Online Adversarial Erasing (OAE) step to enforce two classification network of the same architecture to locate different but complementary regions of target actions.
	
	In the first stage, the dilated convolution module handles the video feature sequence as input to generate the localization sequence $ \mathbf{H} $. Then we conduct a threshold on $ \mathbf{H} $ to generate a mask which represents the discriminative regions detected by the first classifier. Next we use this mask to erase the input video feature maps which are then fed to the second stage. In this way, the second classifier with the erased input feature maps would be forced to discover other action-related regions for supporting the video-level class labels. Concretely, the second classifier would generate new initial seed and then use dilated convolution to expand it. Finally, we integrate the two generated Temporal Class Activation Sequence (T-CAS), $ \mathbf{H} $ and $ \bar{\mathbf{H}} $, to form the cascaded localization sequence $\mathbf{H}_{t}^{k}(Cas)$ = $max\{\mathbf{H}_{t}^{k}, \bar{\mathbf{H}}_{t}^{k}\}$, where $ \mathbf{H}_{t}^{k}(Cas) $ indicates the $ t$-th element in the cascaded localization sequence of class $ k $.
	
	\setlength{\tabcolsep}{2pt}
	\begin{figure}
		\setlength{\abovecaptionskip}{-0.4cm} 
		\begin{center}
			\begin{minipage}[b]{1.0\linewidth}
				\centering
				\centerline{\includegraphics[height=1.6cm]{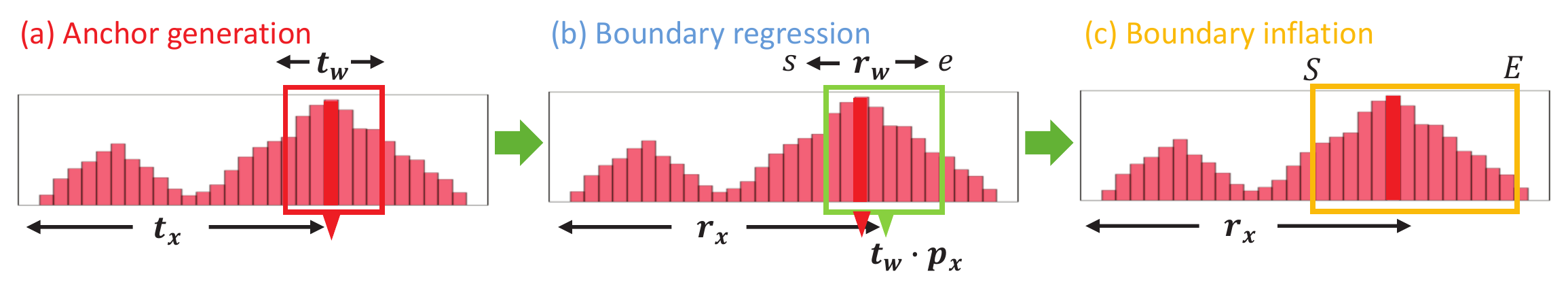}}
				\medskip
			\end{minipage}
		\end{center}
		\caption{Illustration of the boundary prediction procedure which consists of three steps sequentially: (a) \textbf{anchor generation} to obtain the predefined boundary hypothesis; (b) \textbf{boundary regression} to obtain the predicted boundary of the action segment (denoted as the inner boundary); (c) \textbf{boundary inflation} to obtain the outer boundary used for OIC loss implementations.}
		\label{brm}
	\end{figure}
	
\noindent	{\bf Boundary Regression Module.} The goal of this module is to learn to directly predict the segment boundary based on the enhanced CAS obtained above. Multi-anchor mechanism has shown great effectiveness in fully-supervised temporal action detection task, which generates the detections through constantly regressing the pre-defined multi-scale anchors at each temporal position, including the center location and the temporal length respectively. However, without temporal annotations for weakly supervised counterpart,  it is vital to leverage other priors for providing a segment-level supervision. Based on the idea in \cite{Shou2018AutoLoc}, we employ the Outer-Inner-Contrastive (OIC) loss to optimize the predictor. The illustration of the boundary prediction procedure is shown in Fig. \ref{brm}.
	
	Specifically, given the encoded video features as input, a boundary predictor firstly stacks three same temporal convolutional layers to handle the temporal information. Each temporal convolutional layer has the same configurations: 128 filters, kernel size 3, and stride 1 with ReLU activation. Then another temporal convolutional layer with $ 2M $ filters, kernel size 3 and stride 1 is employed to predict the boundary regression values $ p_{x} $ and $ p_{w} $ for each position, where $ M $ is the number of anchor scales. As for anchor generation, we denote $ t_{x} $ and $ t_{w} $ as temporal position and temporal length of each anchor at the output feature maps, and each cell of output feature maps in the prediction layer is associated with multi-scale anchors. Then for each anchor at location $ t_{x} $, we use the output regression values to adjust the segment, where the center localization is $ r_{x} = t_{x} + t_{w} \cdot p_{x}$ and the temporal length is $ r_{w} = t_{w}\cdot exp(p_{w})$, hence the predicted inner boundary can be computed by $ s = r_{x} - r_{w}/2 $ and $ e = r_{x} + r_{w}/2$. Afterwards, in order to implement the OIC loss, we inflate the inner boundary by a ratio $ \gamma $ to obtain the outer boundary $ S = s - r_{w} \cdot \gamma$ and $ E = e + r_{w} \cdot \gamma $. 
	
	OIC loss is introduced to measure how likely the anchor covers the actions and then discard the negative segments. Concretely, it can be denoted as the average activations of the outer red area minus the average activations of the inner green area among the enhanced CAS obtained before.

	\subsection{Transfer Learning Mechanism}
	\label{ssec:am}
	
	Transfer learning technology has been widely explored in modeling the shifts of data distributions across different domains \cite{yan2019cross}, \cite{jing2016predicting}. The performance of video encoders pre-trained on trimmed videos is bound to decrease on untrimmed videos due to great amount of background noises, which influences the quality of CAS a lot. In order to promote the classification performance of untrimmed videos and make full use of the large-scale trimmed video datasets, we introduce the transfer learning mechanism to learn transferable knowledge between trimmed videos and untrimmed videos. Since the trimmed videos are precisely annotated with action of interest, decisive clues in high layers ($ i.e. $ classification layer) can be utilized for action recognition even in untrimmed videos. As shown in Fig.\ref{pic:MGFN}, we take trimmed branch as source branch and untrimmed branch as target branch, then we mine informative knowledge from the trimmed videos to improve the performance on untrimmed videos via knowledge transfer.
	
	For the trimmed branch, we use the identical setting to the untrimmed one. Specifically, trimmed videos are also fed to the two-stream network for feature extraction, then visual feature sequence serves as input of our framework. Different from the untrimmed branch, since the trimmed videos are well-segmented, the boundary regression module is no longer necessary. As a result, the trimmed branch is trained by minimizing the classification loss on the trimmed video dataset as the same as the untrimmed branch.
	After the trimmed branch is convergent, we extract the output features in the GAP layer of classification stage as decisive knowledge fed to the knowledge transfer module. Then we utilize the Maximum Mean Discrepancy (MMD) \cite{Gretton2012A} to measure the distance of the two branches output distribution. In this way, instructive clue is leveraged from trimmed branch to untrimmed branch aiming at improving the overall performance.
	
For more details about the integrated training and inference strategies of KT-MGFN, please refer to our previous published work~\cite{su2020transferable}.

 \section{Experiments}
\subsection{Feature Encoding}

In order to maximize the quality and diversity of video features, we adopt many other ConvNet architectures pre-trained on Kinetics-700 dataset to further fine-tune on HACS dataset (Segments and Clips) with both modalities, including ResNet-50, BN-Inception, SlowFast-101~\cite{feichtenhofer2019slowfast}, pseudo-3D network~\cite{qiu2017learning}, TIN~\cite{shao2020temporal}, TSM~\cite{lin2019tsm} and ResNest-269~\cite{zhang2020resnest}, which are then adopted as visual encoders for feature extraction. Table \ref{feature} illustrates the exact list of two-stream encoders which we adopt to extract video representations. \textbf{Note that we don't use HACS Clips dataset for the supervised learning track}.

For a given video, the frame rate of videos is set to $r$ fps, and each snippet contains $s$ frames. Each snippet is encoded into a visual feature $f_i\in R^C$ by a feature extractor. Given an untrimmed video, a video feature sequence of $F={\{f_i\}}_{i=1}^T\in R^{T\times C}$ is obtained by this method, where $T=l/\delta$, $l$ is the total number of video frames, and $\delta$ is the number of frames interval between different snippets.

 To fuse these features for enriching results diversity, we train our method with these encoders separately, and then fuse the output results through weighted combination instead of averaging, where the normalized weights are calculated according to the relative performance of each feature.

  \setlength{\tabcolsep}{16pt}
  \begin{table*}[t]
  	\centering
  	\caption{List of two-stream encoders with different backbones used to extract video representations.}
  	\small
  	\begin{tabular}{p{1cm}p{4cm}<{\centering}p{4cm}<{\centering}}
  		\toprule
  		\textbf{ID} & \textbf{RGB Feature} & \textbf{Flow Feature}   \\
	 	\hline
  		1 & TSM\_rgb\_ResNest269\_K600 & TSM\_flow\_ResNest269\_K600  \\
		2 & TSN\_rgb\_BnInception\_K600 & TSN\_flow\_BnInception\_K600 \\
		3 & Slowfast101\_rgb\_K700 & Slowfast101\_flow\_K700 \\
		4 & TIN\_rgb\_ResNet101\_K700 & TIN\_flow\_ResNet101\_K700 \\
		5 & P3D\_rgb\_ResNet152\_K600 & P3D\_flow\_ResNet152\_K400 \\
  		\bottomrule
  	\end{tabular}
  	\label{feature}
  \end{table*}

\setlength{\tabcolsep}{15pt}
\begin{table*}[t]
	\centering
	\caption{Action detection results on validation and testing set of HACS in terms of average mAP.}
	\small
	\begin{tabular}{p{4cm}<{\centering}p{0.9cm}<{\centering}p{0.9cm}<{\centering}}
		\toprule
		\multicolumn{3}{c}{ {\bf HACS}, mAP@$tIoU$}  \\
		\hline
		& \multicolumn{1}{c}{validation} & \multicolumn{1}{c}{testing} \\
		\hline
		Method  &  Average  & Average \\
		\hline
		Baseline (Simple CAS) & 15.3  & - \\
		\textbf{+ MDCM} & 20.7 & - \\
		\textbf{+  CCM} & 22.6 &  - \\
		\textbf{+ KTM} & 25.2 & - \\
		\textbf{+  BRM (KT-MGFN)} & \textbf{28.62}  & \textbf{29.78} \\
		\bottomrule
	\end{tabular}
	\label{table_detection_anet}
\end{table*}

\subsection{Supervised Learning Track}
	With our proposed TCANet, we can obtain \textbf{39.91} average mAP on the challenge testing set of supervised learning track with only single model.

\subsection{Weakly-Supervised Learning Track}
	The ablational detection performance comparisons of our method are shown in Table \ref{table_detection_anet}. We can observe that our KT-MGFN can achieve convincing performance. Note that each improvement is performed upon the last one.
 	Finally, our ensembled method achieves \textbf{28.62} average mAP on the validation set of HACS dataset and \textbf{29.78} average mAP on the testing server of weakly-supervised track.

	\section{Conclusion}
	In this challenge notebook, we exhaustively explore the state-of-the-art methods on both supervised and weakly-supervised temporal action localization tasks. For the supervised learning track, we adopt a Temporal Context Aggregation Network (TCANet) for temporal action proposal refinement. Specifically, we first introduce the Local-Global Temporal Encoder (LGTE) to capture both \textit{local and global} temporal relationships simultaneously in a channel grouping fashion. Then the complementary boundary regression mechanism is designed to obtain more precise boundaries and confidence scores.	For the weakly supervised learning track, we propose a unified network named as transferable knowledge based Multi-Granularity Fusion Network (KT-MGFN) for WSTAL. Our method can generate the high-quality Class Activation Sequence (CAS) via augmenting the simple classification network with the cascaded dilated convolution block, where multi-dilated convolution module employs convolutional kernels with varying dilation rates for local discriminative regions expanding, while cascaded classification module adapts two cascaded classifiers for entire regions mining. Besides, in order to improve the classification performance on untrimmed videos, informative knowledge has been transferred from trimmed videos to untrimmed videos with a knowledge transfer module. Finally, a boundary regression module is adopted to perform boundary prediction on the enhanced CAS.

	{\small
		\bibliographystyle{ieee}
		\bibliography{egbib}

\begin{thebibliography}{10}\itemsep=-1pt

\bibitem{caba2015activitynet}
F.~Caba~Heilbron, V.~Escorcia, B.~Ghanem, and J.~Carlos~Niebles.
\newblock Activitynet: A large-scale video benchmark for human activity
  understanding.
\newblock In {\em Proceedings of the ieee conference on computer vision and
  pattern recognition}, pages 961--970, 2015.

\bibitem{chao2018tal_net}
Y.-W. Chao, S.~Vijayanarasimhan, B.~Seybold, D.~A. Ross, J.~Deng, and
  R.~Sukthankar.
\newblock Rethinking the faster r-cnn architecture for temporal action
  localization.
\newblock In {\em Proceedings of the IEEE Conference on Computer Vision and
  Pattern Recognition}, pages 1130--1139, 2018.

\bibitem{feichtenhofer2019slowfast}
C.~Feichtenhofer, H.~Fan, J.~Malik, and K.~He.
\newblock Slowfast networks for video recognition.
\newblock In {\em Proceedings of the IEEE International Conference on Computer
  Vision}, pages 6202--6211, 2019.

\bibitem{gao2018ctap}
J.~Gao, K.~Chen, and R.~Nevatia.
\newblock Ctap: Complementary temporal action proposal generation.
\newblock In {\em Proceedings of the European conference on computer vision
  (ECCV)}, pages 68--83, 2018.

\bibitem{gao2020rapnet}
J.~Gao, Z.~Shi, G.~Wang, J.~Li, Y.~Yuan, S.~Ge, and X.~Zhou.
\newblock Accurate temporal action proposal generation with relation-aware
  pyramid network.
\newblock In {\em AAAI}, pages 10810--10817, 2020.

\bibitem{CBR}
J.~Gao, Z.~Yang, and R.~Nevatia.
\newblock Cascaded boundary regression for temporal action detection.
\newblock {\em arXiv preprint arXiv:1705.01180}, 2017.

\bibitem{gao2017turn}
J.~Gao, Z.~Yang, C.~Sun, K.~Chen, and R.~Nevatia.
\newblock Turn tap: Temporal unit regression network for temporal action
  proposals.
\newblock In {\em ICCV}, pages 3648--3656. IEEE, 2017.

\bibitem{Gretton2012A}
A.~Gretton, K.~M. Borgwardt, M.~Rasch, B.~Schölkopf, and A.~Smola.
\newblock A kernel two-sample test.
\newblock {\em Journal of Machine Learning Research}, 13(1):723--773, 2012.

\bibitem{Guo2018Fully}
D.~Guo, L.~Wei, and X.~Fang.
\newblock Fully convolutional network for multiscale temporal action proposals.
\newblock {\em IEEE Transactions on Multimedia}, 20(12):3428--3438, 2018.

\bibitem{Hao2018Temporal}
S.~Hao, X.~Wu, Z.~Bing, Y.~Wu, and Y.~Jia.
\newblock Temporal action localization in untrimmed videos using action pattern
  trees.
\newblock {\em IEEE Transactions on Multimedia}, PP(99):1--1, 2018.

\bibitem{thumos14}
Y.~Jiang, J.~Liu, A.~R. Zamir, G.~Toderici, I.~Laptev, M.~Shah, and
  R.~Sukthankar.
\newblock Thumos challenge: Action recognition with a large number of classes.
\newblock In {\em Computer Vision-ECCV workshop 2014}, 2014.

\bibitem{jing2016predicting}
P.~Jing, Y.~Su, L.~Nie, and H.~Gu.
\newblock Predicting image memorability through adaptive transfer learning from
  external sources.
\newblock {\em IEEE Transactions on Multimedia}, 19(5):1050--1062, 2016.

\bibitem{lin2019tsm}
J.~Lin, C.~Gan, and S.~Han.
\newblock Tsm: Temporal shift module for efficient video understanding.
\newblock In {\em Proceedings of the IEEE International Conference on Computer
  Vision}, pages 7083--7093, 2019.

\bibitem{lin2019bmn}
T.~Lin, X.~Liu, X.~Li, E.~Ding, and S.~Wen.
\newblock Bmn: Boundary-matching network for temporal action proposal
  generation.
\newblock In {\em Proceedings of the IEEE International Conference on Computer
  Vision}, pages 3889--3898, 2019.

\bibitem{LinBMN}
T.~Lin, X.~Liu, X.~Li, E.~Ding, and S.~Wen.
\newblock Bmn: Boundary-matching network for temporal action proposal
  generation.
\newblock {\em CoRR abs/1907.09702}, 2019.

\bibitem{SSAD}
T.~Lin, X.~Zhao, and Z.~Shou.
\newblock Single shot temporal action detection.
\newblock In {\em Proceedings of the 2017 ACM on Multimedia Conference}, pages
  988--996. ACM, 2017.

\bibitem{lin2017ssad}
T.~Lin, X.~Zhao, and Z.~Shou.
\newblock Single shot temporal action detection.
\newblock In {\em Proceedings of the 25th ACM international conference on
  Multimedia}, pages 988--996, 2017.

\bibitem{BSN}
T.~Lin, X.~Zhao, H.~Su, C.~Wang, and M.~Yang.
\newblock Bsn: Boundary sensitive network for temporal action proposal
  generation.
\newblock {\em arXiv preprint arXiv:1806.02964}, 2018.

\bibitem{lin2018bsn}
T.~Lin, X.~Zhao, H.~Su, C.~Wang, and M.~Yang.
\newblock Bsn: Boundary sensitive network for temporal action proposal
  generation.
\newblock In {\em Proceedings of the European Conference on Computer Vision
  (ECCV)}, pages 3--19, 2018.

\bibitem{liu2019mgg}
Y.~Liu, L.~Ma, Y.~Zhang, W.~Liu, and S.-F. Chang.
\newblock Multi-granularity generator for temporal action proposal.
\newblock In {\em Proceedings of the IEEE Conference on Computer Vision and
  Pattern Recognition}, pages 3604--3613, 2019.

\bibitem{qing2021temporal}
Z.~Qing, H.~Su, W.~Gan, D.~Wang, W.~Wu, X.~Wang, Y.~Qiao, J.~Yan, C.~Gao, and
  N.~Sang.
\newblock Temporal context aggregation network for temporal action proposal
  refinement.
\newblock {\em arXiv preprint arXiv:2103.13141}, 2021.

\bibitem{qiu2017learning}
Z.~Qiu, T.~Yao, and T.~Mei.
\newblock Learning spatio-temporal representation with pseudo-3d residual
  networks.
\newblock In {\em proceedings of the IEEE International Conference on Computer
  Vision}, pages 5533--5541, 2017.

\bibitem{shao2020temporal}
H.~Shao, S.~Qian, and Y.~Liu.
\newblock Temporal interlacing network.
\newblock {\em AAAI}, 2020.

\bibitem{CDC}
Z.~Shou, J.~Chan, A.~Zareian, K.~Miyazawa, and S.-F. Chang.
\newblock Cdc: Convolutional-de-convolutional networks for precise temporal
  action localization in untrimmed videos.
\newblock In {\em CVPR}, pages 1417--1426. IEEE, 2017.

\bibitem{Shou2018AutoLoc}
Z.~Shou, H.~Gao, L.~Zhang, K.~Miyazawa, and S.-F. Chang.
\newblock Autoloc: Weakly-supervised temporal action localization in untrimmed
  videos.
\newblock In {\em European Conference on Computer Vision}, 2018.

\bibitem{SCNN}
Z.~Shou, D.~Wang, and S.-F. Chang.
\newblock Temporal action localization in untrimmed videos via multi-stage
  cnns.
\newblock In {\em CVPR}, pages 1049--1058, 2016.

\bibitem{shou2016scnn}
Z.~Shou, D.~Wang, and S.-F. Chang.
\newblock Temporal action localization in untrimmed videos via multi-stage
  cnns.
\newblock In {\em Proceedings of the IEEE Conference on Computer Vision and
  Pattern Recognition}, pages 1049--1058, 2016.

\bibitem{singh2016anet_winner}
G.~Singh and F.~Cuzzolin.
\newblock Untrimmed video classification for activity detection: submission to
  activitynet challenge.
\newblock {\em arXiv preprint arXiv:1607.01979}, 2016.

\bibitem{su2020transferable}
H.~Su, X.~Zhao, T.~Lin, S.~Liu, and Z.~Hu.
\newblock Transferable knowledge-based multi-granularity fusion network for
  weakly supervised temporal action detection.
\newblock {\em IEEE Transactions on Multimedia}, 2020.

\bibitem{vaswani2017transformer}
A.~Vaswani, N.~Shazeer, N.~Parmar, J.~Uszkoreit, L.~Jones, A.~N. Gomez,
  {\L}.~Kaiser, and I.~Polosukhin.
\newblock Attention is all you need.
\newblock In {\em Advances in neural information processing systems}, pages
  5998--6008, 2017.

\bibitem{wang2015ssn}
L.~Wang, Y.~Xiong, Z.~Wang, and Y.~Qiao.
\newblock Towards good practices for very deep two-stream convnets.
\newblock {\em arXiv preprint arXiv:1507.02159}, 2015.

\bibitem{wang2016tsn}
L.~Wang, Y.~Xiong, Z.~Wang, Y.~Qiao, D.~Lin, X.~Tang, and L.~Van~Gool.
\newblock Temporal segment networks: Towards good practices for deep action
  recognition.
\newblock In {\em European conference on computer vision}, pages 20--36.
  Springer, 2016.

\bibitem{wang2018nonlocal}
X.~Wang, R.~Girshick, A.~Gupta, and K.~He.
\newblock Non-local neural networks.
\newblock In {\em Proceedings of the IEEE conference on computer vision and
  pattern recognition}, pages 7794--7803, 2018.

\bibitem{wu2019lfb}
C.-Y. Wu, C.~Feichtenhofer, H.~Fan, K.~He, P.~Krahenbuhl, and R.~Girshick.
\newblock Long-term feature banks for detailed video understanding.
\newblock pages 284--293, 2019.

\bibitem{xiong2016cuhk}
Y.~Xiong, L.~Wang, Z.~Wang, B.~Zhang, H.~Song, W.~Li, D.~Lin, Y.~Qiao,
  L.~Van~Gool, and X.~Tang.
\newblock Cuhk \& ethz \& siat submission to activitynet challenge 2016.
\newblock {\em arXiv preprint arXiv:1608.00797}, 2016.

\bibitem{yan2019cross}
C.~Yan, L.~Li, C.~Zhang, B.~Liu, Y.~Zhang, and Q.~Dai.
\newblock Cross-modality bridging and knowledge transferring for image
  understanding.
\newblock {\em IEEE Transactions on Multimedia}, 2019.

\bibitem{zhang2020resnest}
H.~Zhang, C.~Wu, Z.~Zhang, Y.~Zhu, Z.~Zhang, H.~Lin, Y.~Sun, T.~He, J.~Mueller,
  R.~Manmatha, et~al.
\newblock Resnest: Split-attention networks.
\newblock {\em arXiv preprint arXiv:2004.08955}, 2020.

\bibitem{zhao2019hacs}
H.~Zhao, A.~Torralba, L.~Torresani, and Z.~Yan.
\newblock Hacs: Human action clips and segments dataset for recognition and
  temporal localization.
\newblock In {\em Proceedings of the IEEE International Conference on Computer
  Vision}, pages 8668--8678, 2019.

\bibitem{SSN}
Y.~Zhao, Y.~Xiong, L.~Wang, Z.~Wu, X.~Tang, and D.~Lin.
\newblock Temporal action detection with structured segment networks.
\newblock In {\em ICCV}, volume~2, 2017.

\bibitem{B.Zhou}
B.~Zhou, A.~Khosla, A.~Lapedriza, A.~Oliva, and A.~Torralba.
\newblock Learning deep features for discriminative localization.
\newblock In {\em CVPR}, pages 2921--2929, 2016.

\bibitem{zhou2015learning}
Z.~Zhou, F.~Shi, and W.~Wu.
\newblock Learning spatial and temporal extents of human actions for action
  detection.
\newblock {\em IEEE Transactions on Multimedia}, 17(4):512--525, 2015.

\end{thebibliography}
	}
	
\end{document}